\documentclass[10pt,twocolumn,letterpaper]{article}

\usepackage{iccv}
\usepackage{times}
\usepackage{epsfig}
\usepackage{graphicx}
\usepackage{amsmath}
\usepackage{amssymb}
\usepackage{multirow}
\usepackage{booktabs}
\usepackage{threeparttable}
\usepackage{subfig}
\usepackage{authblk}

\usepackage[breaklinks=true,bookmarks=false]{hyperref}

\iccvfinalcopy 

\begin{document}

\title{Progressive Differentiable Architecture Search: \\
Bridging the Depth Gap between Search and Evaluation}

\author{Xin Chen$^{1}$\thanks{This work was done when the first author was an intern at Huawei
Noah’s Ark Lab.} \qquad Lingxi Xie$^{2}$\qquad Jun Wu$^{1}$ \qquad Qi Tian$^{2}$\\
$^{1}$Tongji University \quad $^{2}$Huawei Noah's Ark Lab\\
{\tt\small 1410452@tongji.edu.cn \qquad 198808xc@gmail.com \qquad wujun@tongji.edu.cn \qquad tian.qi1@huawei.com}}

\maketitle

\begin{abstract}
Recently, differentiable search methods have made major progress in reducing the computational costs of neural architecture search.
However, these approaches often report lower accuracy in evaluating the searched architecture or transferring it to another dataset.
This is arguably due to the large gap between the architecture depths in search and evaluation scenarios.
In this paper, we present an efficient algorithm which allows the depth of searched architectures to grow gradually during the training procedure.
This brings two issues, namely, heavier computational overheads and weaker search stability,
which we solve using search space approximation and regularization, respectively.
With a significantly reduced search time ($\sim$7 hours on a single GPU), our approach achieves state-of-the-art performance on both the proxy dataset (CIFAR10 or CIFAR100) and the target dataset (ImageNet).
Code is available at \url{https://github.com/chenxin061/pdarts}.
\end{abstract}

\section{Introduction}

Image recognition is a fundamental task in the computer vision community.
In the deep learning era, state-of-the-art classification performance is mostly achieved by handcrafted deep neural networks.
Recently, the development of neural architecture search (NAS) has changed the convention of model design from manual to automatic, achieving remarkable success in various perceptual tasks~\cite{liu2019auto, chu2019fast, zoph2016neural} including image recognition~\cite{zoph2018learning}.

Early works on NAS focused on the optimal configuration of layer type, filter size and number, activation function, {\em etc.}, to construct a complete network~\cite{baker2016designing, suganuma2017genetic}.
Inspired by successful handcrafted architectures such as ResNet~\cite{he2016deep} and DenseNet~\cite{huang2017densely},
follow-up works started to explore the possibility of searching for network building blocks, or so-called cells with reinforcement learning (RL)~\cite{zhong2018practical, zoph2018learning} and evolutionary algorithm (EA)~\cite{xie2017genetic, real2018regularized}.
The discovered cells are then stacked orderly to construct the network for specific tasks.
However, those RL-based and EA-based approaches share a common pipeline to sample and evaluate (from scratch) numerous architectures in the search space,
which results in a barely affordable computational overhead, {\em e.g.},  hundreds or even thousands of GPU-days.

\begin{figure}[t]
\begin{center}
   \includegraphics[width=0.99\linewidth]{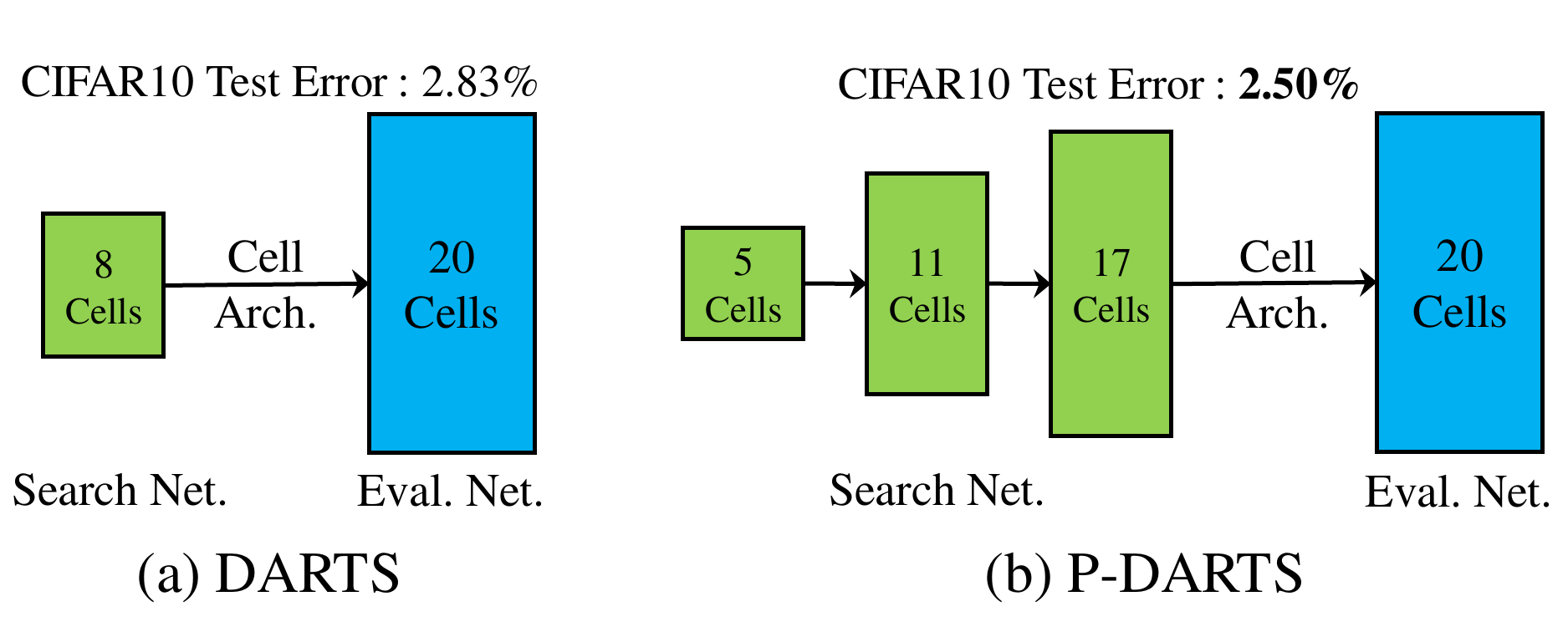}
\end{center}
\caption{Difference between DARTS and P-DARTS (our approach), with the former searching architectures in a shallow setting and evaluating them in a deep one, and the latter progressively increasing the searching depth, so as to bridge the {\em depth gap} between search and evaluation. Green and blue indicate search and evaluation, respectively.}
\label{motivation}
\end{figure}

Recently, Liu {\em et al.} proposed a differentiable scheme called DARTS~\cite{liu2018darts} to get rid of the time-consuming process of architecture sampling and evaluating.
It achieved comparable performance to RL-based and EA-based methods while only requiring a search cost of a few GPU-days.
In DARTS, a cell is composed of multiple nodes, connected with several kinds of operations, {\em e.g.}, {\it convolution}, {\it pooling}.
Those operations are weighted by a few architecture parameters,
which are learned in the search scenario.
Limited by the size of GPU memory, DARTS has to search the architecture in a shallow network while evaluate in a deeper one.
This brings an issue named the {\em depth gap} (see Figure~\ref{motivation}(a)), which means that the search stage finds some operations that work well in a shallow architecture, but the evaluation stage actually prefers other operations that fit a deep architecture better.
Such gap hinders these approaches in their application to more complex visual recognition tasks.

In this work, we propose Progressive DARTS (P-DARTS), a novel and efficient algorithm to bridge the depth gap.
As shown in Figure~\ref{motivation}(b), we divide the search process into multiple stages and progressively increase the network depth at the end of each stage.
While a deeper architecture requires heavier computational overhead, we propose {\em search space approximation} which, as the depth increases, reduces the number of candidates (operations) according to their scores in the elapsed search process.
Another issue, lack of stability, emerges with searching over a deep architecture,
in which the algorithm can be biased heavily towards {\it skip-connect} as it often leads to rapidest error decay during optimization,
but, actually, a better option often resides in learnable operations such as {\it convolution}.
To avoid this, we propose {\em search space regularization},
which (i) introduces operation-level Dropout~\cite{srivastava2014dropout} to alleviate the dominance of {\it skip-connect} during training,
and (ii) controls the appearance of {\it skip-connect} during evaluation.

The effectiveness of P-DARTS is verified on the standard vision setting,
{\em i.e.}, searching on CIFAR10, and evaluating on both CIFAR10 and ImageNet.
We achieve state-of-the-art performance (a test error of 2.50\%) on CIFAR10 with 3.4M parameters.
When transferred to ImageNet, it achieves top-1/5 errors of 24.4\%/7.4\%, respectively,
comparable to the state-of-the-art under the mobile setting.
We further demonstrate the benefits of search space approximation and regularization:
the former reduces the search time over CIFAR10 to 0.3 GPU-days which, to the best of our knowledge, is the fastest to date to achieve an error rate of 3\% in CIFAR10, even surpassing ENAS~\cite{pham2018efficient},
an approach specialized in efficiency;
the latter makes it easy to apply P-DARTS to other proxy datasets,
which we show an example on CIFAR100 (15.92\% test error, 3.6M parameters).

\begin{figure*}[t]
\begin{center}
   \includegraphics[width=0.95\linewidth]{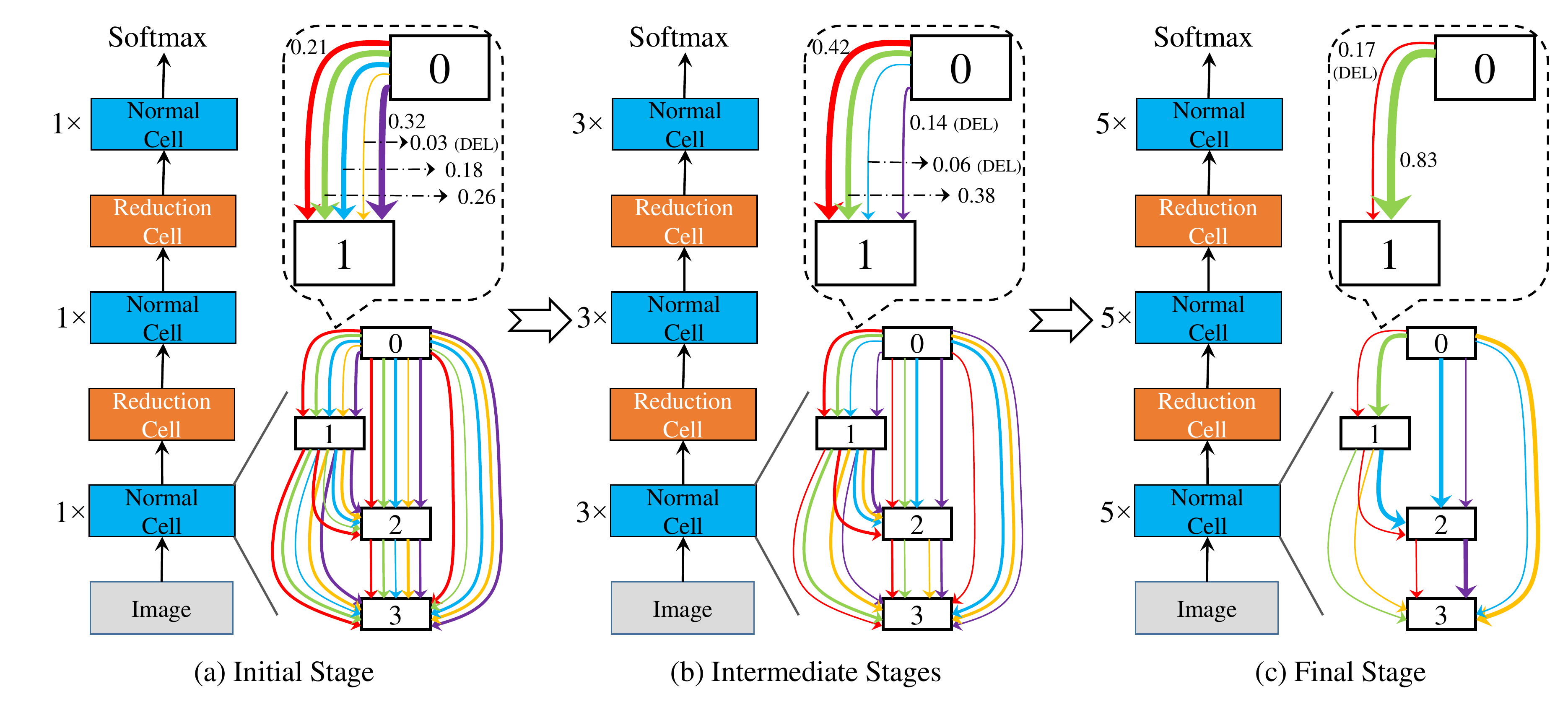}
\end{center}
\caption{The overall pipeline of P-DARTS (best viewed in color). For simplicity, only one intermediate stage is shown, and only the normal cells are displayed. The depth of the search network increases from 5 at the initial stage to 11 and 17 at the intermediate and final stages, while the number of candidate operations (shown in connections with different colors) is shrunk from 5 to 3 and 2 accordingly. The lowest-scored ones at the previous stage are dropped (the scores are shown next to each connection). We obtain the final architecture by considering the final scores and possibly additional rules.}
\label{pipeline}
\end{figure*}

\section{Related Work}

Image recognition is a fundamental task in computer vision.
Recent years, with the development of deep learning, convolutional neural networks (CNNs) have been dominating image recognition~\cite{krizhevsky2012imagenet}.
A few handcrafted architectures have been proposed, including VGGNet~\cite{simonyan2014very}, ResNet~\cite{he2016deep}, DenseNet~\cite{huang2017densely}, {\em etc.}, all of which verified the importance of human experts in network design.

Our work belongs to the emerging field of neural architecture search (NAS), a process of automating architecture engineering technique~\cite{elsken2018neural}.
Pioneer researchers started to explore the possibility of automatically generating better topology with evolutionary algorithms in the 2000's~\cite{stanley2002evolving}.
Early NAS works tried to search for a complete network topology~\cite{baker2016designing, suganuma2017genetic} while recent works focused on finding robust cells~\cite{xie2017genetic, zoph2018learning, real2018regularized}.
Lately, EA-based~\cite{real2018regularized} and RL-based~\cite{zoph2018learning} NAS approaches achieved state-of-the-art performance in image recognition,
where architectures were sampled and evaluated from the search space under the guidance of an EA-based or RL-based meta-controller.
A notable drawback of the above approaches is the expensive computational overhead (3,150 GPU-days for EA-based AmoebaNet~\cite{real2018regularized} and 1,800 GPU-days for RL-based NASNet~\cite{zoph2018learning}).
PNAS proposed to learn a surrogate model to guide the search through the structure space, achieving 5$\times$ speedup than NASNet.
ENAS~\cite{pham2018efficient} proposed to share parameters among child models to prevent evaluating candidate architectures by training them from scratch, which significantly reduced the search cost to less than one GPU-day.

DARTS~\cite{liu2018darts} introduced a differentiable NAS framework, which achieved remarkable performance and efficiency improvement.
Following DARTS, SNAS~\cite{xie2018snas} proposed to constrain the architecture parameters to be one-hot to tackle the inconsistency in optimizing objectives between search and evaluation scenarios.
ProxylessNAS~\cite{cai2018proxylessnas} adopted the differentiable framework and proposed to search architectures on the target task instead of adopting the conventional proxy-based framework.

\section{Method}

\subsection{Preliminary: DARTS}

In this work, we leverage DARTS~\cite{liu2018darts} as our baseline framework.
Our goal is to search for a robust cell and apply it to a network of $L$ cells.
A cell is defined as a directed acyclic graph (DAG) of $N$ nodes, $\{x_0, x_1, \cdots, x_{N-1}\}$, where each node is a network layer, {\em i.e.}, performing a specific mathematical function.
We denote the operation space as $\mathcal{O}$, in which each element represents a candidate function $o(\cdot)$.
An edge $\mathrm{E}_{(i, j)}$ represents the information flow connecting node $i$ and node $j$, which consists of a set of operations weighted by the architecture parameters $\alpha^{(i, j)}$, and is thus formulated as:
\begin{equation}
  f_{i, j}(x_i) = \sum_{o\in\mathcal{O}_{i, j}}{\frac{\mathrm{exp}(\alpha_o^{(i,j)})}{\sum_{o'\in\mathcal{O}}\mathrm{exp}(\alpha_{o'}^{(i,j)})}o(x_i)},
\end{equation}
where $i<j$ so that {\it skip-connect} can be applied.
An intermediate node can be represented as $x_j = \sum_{i < j}{f_{i,j}(x_i)}$,
and the output node is $x_{N-1} = \mathrm{concat}(x_2, x_3, \cdots, x_{N-2})$,
where $\mathrm{concat}(\cdot)$ concatenates all input signals in the channel dimension.
For more technical details, please refer to the original DARTS paper~\cite{liu2018darts}.

\subsection{Progressively Increasing the Searching Depth}

In DARTS, architecture search is performed on a network of 8 cells while the discovered architecture is evaluated on a network of 20 cells.
However, there is a big difference between the behaviors of shallow and deep networks~\cite{ioffe2015batch, srivastava2015training, he2016deep}, which implies that the structures we prefer in the search process are not necessarily the optimal for evaluation.
We name this the {\bf depth gap} between search and evaluation.
To verify it, we executed the search process of DARTS for multiple times and found that the normal cells of discovered architectures tend to keep shallow connections instead of deep ones.
This is caused by that shallow networks often enjoy faster gradient descent during the search process, but contradicts the common sense that deeper networks tend to perform better~\cite{simonyan2014very,szegedy2015going,he2016deep,huang2017densely}. Therefore, we propose to bridge the depth gap, and we take the strategy that progressively increases the network depth during the search process,
so that at the end of search, the depth is sufficiently close to the setting used in evaluation.
Here we prefer a progressive manner, rather than directly increasing the depth to the target level, because we expect search in shallow networks to reduce the search space with respect to the candidate operations, so as to alleviate the risk of search in deep networks.
We will verify the effectiveness of this progressive strategy in Section~\ref{comp_depth}.

Difficulty comes from two aspects.
First, the number of structures increases exponentially with the depth, which brings issues in both time and memory.
In particular, in DARTS, GPU memory usage is proportional to the depth of searched networks.
The limited GPU memory forms a major obstacle, and the most straightforward solution is to reduce the number of channels in each operation -- DARTS~\cite{liu2018darts} tried it but reported a slight performance deterioration.
To address this problem, we propose a {\bf search space approximation} scheme to progressively reduce the number of candidate operations at the end of each stage, which refers to the scores of operations in the previous stage as the criterion of selection.
Details of search space approximation are presented in Section~\ref{ssa}.

Second, we find that when searching on a deeper architecture, the differentiable approaches tend to bias towards the {\it skip-connect} operation,
because it accelerates forward/backward propagation and often leads to the fastest way of gradient descent.
However, since such an operation is parameter-free, its ability of learning visual representations is relatively weak.
To this end, we propose another scheme named {\bf search space regularization}, which adds operation-level Dropout~\cite{srivastava2014dropout} to prevent the architecture from `over-fitting' and restricts the number of preserved {\it skip-connects} for further stability.
Details of search space regularization are presented in Section~\ref{ssr}.

\subsubsection{Search Space Approximation}
\label{ssa}

The idea of search space approximation is shown as the toy example in Figure~\ref{pipeline}.
The search process is split into multiple stages, including an initial stage, one or a few intermediate stages and a final stage.
For each stage, $\mathfrak{S}_k$, the search network consists of $L_k$ cells and the size of the operation space is $O_k$, {\em i.e.},  $|\mathcal{O}_{(i, j)}^k| = O_k$.

According to our motivation, at the initial stage, the search network is relatively shallow but the operation space is large ($\mathcal{O}_{(i, j)}^1 \equiv \mathcal{O}$).
After each stage, $\mathfrak{S}_{k-1}$, the architecture parameters $\alpha_{k-1}$ are learned and the scores of the candidate operations on each connection are ranked according to $\alpha_{k-1}$.
We increase the depth of the searched architecture by stacking more cells, {\em i.e.}, $L_{k}>L_{k-1}$, and approximate the operation space in the meantime.
This is to say, the new operation set $\mathcal{O}_{(i, j)}^k$ has a smaller size than $\mathcal{O}_{(i, j)}^{k-1}$, or equivalently, $O_{k}<O_{k-1}$.
The criterion of approximation is to drop a part of less important operations, which are defined to be those assigned with a lower weight during the previous stage, $\mathfrak{S}_{k-1}$.
As shown in Table~\ref{mem_usage}, this strategy is memory efficient, which makes our approach easy to be deployed on regular GPUs, {\em e.g.}, with a memory of 16GB.

This process of increasing architecture depth continues until it is sufficiently close to that used in evaluation.
After the last search stage, we determine the final cell topology (bold lines in Figure~\ref{pipeline}(c)) according to the learned architecture parameters $\alpha_K$.
Following DARTS, we keep two top-weighted non-{\it zero} operations (at most 1 for a distinct edge) for each intermediate node.

\begin{table*}[t]
\begin{center}
\begin{tabular}{lccccc}
\hline
\textbf{\multirow{2}{*}{Architecture}} & \multicolumn{2}{c}{\textbf{Test Err. (\%})} & \textbf{Params} & \textbf{Search Cost} & \textbf{\multirow{2}{*}{Search Method}} \\
\cmidrule(lr){2-3}
&                            \textbf{C10} & \textbf{C100} & \textbf{(M)} & \textbf{(GPU-days)} &\\
\hline
DenseNet-BC~\cite{huang2017densely}                       & 3.46 & 17.18 & 25.6 & -    & manual \\
\hline
NASNet-A + cutout~\cite{zoph2018learning}                 & 2.65 & -     & 3.3  & 1800 & RL      \\
AmoebaNet-A + cutout~\cite{real2018regularized}           & 3.34 & -     & 3.2  & 3150 & evolution \\
AmoebaNet-B + cutout~\cite{real2018regularized}           & 2.55 & -     & 2.8  & 3150 & evolution \\
Hireachical Evolution~\cite{liu2017hierarchical}          & 3.75 & -     & 15.7 & 300  & evolution \\
PNAS~\cite{liu2018progressive}                            & 3.41 & -     & 3.2  & 225  & SMBO \\
ENAS + cutout~\cite{pham2018efficient}                    & 2.89 & -     & 4.6  & 0.5  & RL \\
\hline
DARTS (first order) + cutout~\cite{liu2018darts}           & 3.00 & 17.76$^\dagger$ & 3.3 & 1.5$^\ddagger$  & gradient-based \\
DARTS (second order) + cutout~\cite{liu2018darts}          & 2.76 & 17.54$^\dagger$ & 3.3 & 4.0$^\ddagger$ & gradient-based \\
SNAS + mild constraint + cutout~\cite{xie2018snas}        & 2.98 & -     & 2.9  & 1.5  & gradient-based \\
SNAS + moderate constraint + cutout~\cite{xie2018snas}    & 2.85 & -     & 2.8  & 1.5  & gradient-based \\
SNAS + aggressive constraint + cutout~\cite{xie2018snas}  & 3.10 & -     & 2.3  & 1.5  & gradient-based \\
ProxylessNAS~\cite{cai2018proxylessnas} + cutout   & 2.08 & -     & 5.7  & 4.0    & gradient-based \\
\hline
P-DARTS CIFAR10 + cutout                                    & 2.50 & 16.55 & 3.4  & 0.3 & gradient-based \\
P-DARTS CIFAR100 + cutout                                   & 2.62 & 15.92 & 3.6  & 0.3 & gradient-based \\
P-DARTS CIFAR10 (large) + cutout                            & 2.25 & 15.27 & 10.5 & 0.3 & gradient-based \\
P-DARTS CIFAR100 (large) + cutout                           & 2.43 & 14.64 & 11.0 & 0.3 & gradient-based \\
\hline
\end{tabular}
\end{center}
\caption{Comparison with state-of-the-art architectures on CIFAR10 and CIFAR100.
$^\dagger$ indicates that this result is obtained by training the corresponding architecture on CIFAR100. $^\ddagger$ We ran the code released by the authors with necessary modifications to fit PyTorch 1.0, and a single run took about 0.5 GPU-days for first order and 2 GPU-days for second order, respectively. }
\label{tab_ev_cifar}
\end{table*}

\subsubsection{Search Space Regularization}
\label{ssr}

At the start of each stage, $\mathfrak{S}_k$, we train the (modified) architecture from scratch, {\em i.e.}, all network weights are initialized, because several candidates have been dropped\footnote{We also tried to start with network parameters learned from the last stage, $\mathfrak{S}_{k-1}$, and adjust the architecture weights accordingly, {\em i.e.}, the weights of preserved operations should still sum to one, and each weight is proportional to the corresponding value at the end of $\mathfrak{S}_{k-1}$. This strategy reported lower accuracy, because the prior from the previous stage guided the algorithm towards an architecture suitable for a shallow network instead of a deep one.}.
However, training a deeper network is harder than training a shallow one~\cite{srivastava2015training}.
In our particular setting, we observe that information prefers to flow through {\it skip-connect} instead of {\it convolution} or {\it pooling}, which is arguably due to the reason that {\it skip-connect} often leads to rapid gradient descent, especially on the proxy datasets (CIFAR10 or CIFAR100) which are relatively small and easy to fit.
Consequently, the search process tends to generate architectures with many {\it skip-connect} operations, which limits the number of learnable parameters and thus produces unsatisfying performance at the evaluation stage.
This is essentially a kind of over-fitting.

We address this problem by search space regularization, which consists of two parts. First, we insert operation-level Dropout~\cite{srivastava2014dropout} after each {\it skip-connect} operation, so as to partially `cut off' the straightforward path through {\it skip-connect}, and facilitate the algorithm to explore other operations.
However, if we constantly block the path through {\it skip-connect}, the algorithm will drop them by assigning low weights to them, which is harmful to the final performance.
To address this contradiction, we gradually decay the Dropout rate during the training process in each search stage, thus the straightforward path through {\it skip-connect} is blocked at the beginning and treated equally afterward when parameters of other operations are well learned, leaving the algorithm itself to make the decision.

Despite the use of Dropout, we still observe that {\it skip-connect}, as a special kind of operation, has a significant impact on recognition accuracy at the evaluation stage.
Empirically, we perform 3 search processes on CIFAR10 with exactly the same search setting, but find that the number of preserved {\it skip-connects} in the normal cell, after the final stage, varies from 2 to 4.
In the meantime, as we observed before, the recognition performance at the evaluation stage is also highly correlated to this number.
This motivates us to design the second regularization rule, architecture refinement, which simply controls the number of preserved {\it skip-connects}, after the final search stage, to be a constant $M$.
This is done with an iterative process, which starts with constructing a cell topology using the standard algorithm described by DARTS.
If the number of {\it skip-connects} is not exactly $M$, we search for the $M$ {\it skip-connect} operations with the largest architecture weights in this cell topology and set the weights of others to 0, then redo cell construction with modified architecture parameters. 
Note that this may bring up other {\it skip-connects} to the topology, so we repeat this procedure until the desired number is achieved.

We emphasize that the second regularization technique must be applied on top of the first one, otherwise, in the situations without operation-level Dropout, the search process is producing low-quality architecture weights, based on which we could not build up a powerful architecture even with a fixed number of {\it skip-connects}.

\subsection{Relationship to Prior Work}

PNAS~\cite{liu2018progressive} explored the search space progressively by searching for operations node-by-node within each cell.
Our approach has a similar search manner but comes from a different motivation.
In addition, we perform the progressive search at the cell level to enlarge the architecture depth, while PNAS did it at the operation level (within a cell) to reduce the number of architectures to evaluate.

SNAS~\cite{xie2018snas} aimed at eliminating the bias between the search and evaluation objectives of differentiable NAS approaches by forcing the architecture weights on each edge to be one-hot.
Our work is also able to get rid of the bias, which we investigate from enlarging the architecture depth.

ProxylessNAS~\cite{cai2018proxylessnas} introduced a differentiable NAS scheme to directly learn architectures on the target task (and hardware) without a proxy dataset.
It achieved high memory efficiency by applying binary masks to operations and forcing only one path in the over-parameterized network to be activated and loaded into GPU.
Different from it, our approach tackles the memory overhead by search space approximation.
Besides, ProxylessNAS searched for global topology instead of cell topology, which requires strong priors on the target task as well as the search space, while P-DARTS does not need such priors.
Our approach is much faster than ProxylessNAS (0.3 GPU-days vs. 4 GPU-days on CIFAR10).

\section{Experiments}\label{exp}

\subsection{Datasets}

We conduct experiments on three popular image classification datasets, including CIFAR10, CIFAR100~\cite{krizhevsky2009learning} and ImageNet~\cite{deng2009imagenet}.
Architecture search is performed on CIFAR10 and CIFAR100, and the discovered architectures are evaluated on all three datasets.

Each of CIFAR10 and CIFAR100 has 50K/10K training/testing RGB images with a fixed spatial resolution of 32$\times$32.
These images are equally distributed over 10/100 classes.
In the architecture search scenario, the training set is equally split into two subsets,
one for tuning network parameters ({\em e.g.}, convolutional weights) and the other for tuning the architecture ({\em i.e.}, operation weights).
In the evaluation scenario, the standard training/testing split is used.

We use ILSVRC2012~\cite{russakovsky2015imagenet} to test the transferability of the architectures discovered on CIFAR10 and CIFAR100.
ILSVRC2012 is a subset of ImageNet~\cite{deng2009imagenet} which contains 1,000 object categories and 1.28M training and 50K validation images.
Following the conventions~\cite{zoph2018learning, liu2018darts}, we apply the mobile setting where the input image size is 224$\times$224 and the number of multi-add operations is restricted to be less than 600M.

\begin{table*} [t]
\begin{center}
\begin{tabular}{lcccccc}
\hline
\textbf{\multirow{2}{*}{Architecture}} & \multicolumn{2}{c}{\textbf{Test Err. (\%)}} & \textbf{Params} & $\times+$ & \textbf{Search Cost} & \textbf{\multirow{2}{*}{Search Method}} \\
\cmidrule(lr){2-3}
&                            \textbf{top-1} & \textbf{top-5} & \textbf{(M)} & \textbf{(M)} & \textbf{(GPU-days)} &\\
\hline
Inception-v1~\cite{szegedy2015going}          & 30.2 & 10.1 & 6.6 & 1448 & -    & manual \\
MobileNet~\cite{howard2017mobilenets}         & 29.4 & 10.5 & 4.2 & 569  & -    & manual \\
ShuffleNet 2$\times$ (v1)~\cite{zhang2018shufflenet} & 26.4 & 10.2 & $\sim$5  & 524  & -    & manual \\
ShuffleNet 2$\times$ (v2)~\cite{ma2018shufflenet}    & 25.1 & - & $\sim$5  & 591  & -    & manual \\
\hline
NASNet-A~\cite{zoph2018learning}              & 26.0 & 8.4  & 5.3 & 564  & 1800 & RL \\
NASNet-B~\cite{zoph2018learning}              & 27.2 & 8.7  & 5.3 & 488  & 1800 & RL \\
NASNet-C~\cite{zoph2018learning}              & 27.5 & 9.0  & 4.9 & 558  & 1800 & RL \\
AmoebaNet-A~\cite{real2018regularized}        & 25.5 & 8.0  & 5.1 & 555  & 3150 & evolution \\
AmoebaNet-B~\cite{real2018regularized}        & 26.0 & 8.5  & 5.3 & 555  & 3150 & evolution \\
AmoebaNet-C~\cite{real2018regularized}        & 24.3 & 7.6  & 6.4 & 570  & 3150 & evolution \\
PNAS~\cite{liu2018progressive}                & 25.8 & 8.1  & 5.1 & 588  & 225  & SMBO \\
MnasNet-92~\cite{tan2018mnasnet}              & 25.2 & 8.0  & 4.4 & 388  & -    & RL \\
\hline
DARTS (second order)~\cite{liu2018darts}      & 26.7 & 8.7  & 4.7 & 574  & 4.0    & gradient-based \\
SNAS (mild constraint)~\cite{xie2018snas}     & 27.3 & 9.2  & 4.3 & 522  & 1.5  & gradient-based \\
ProxylessNAS (GPU)~\cite{cai2018proxylessnas}       & 24.9 & 7.5  & 7.1 & 465  & ~8.3  & gradient-based \\
\hline
P-DARTS (searched on CIFAR10)                 & 24.4 & 7.4  & 4.9 & 557  & 0.3  & gradient-based \\
P-DARTS (searched on CIFAR100)                & 24.7 & 7.5  & 5.1 & 577  & 0.3  & gradient-based \\
\hline
\end{tabular}
\end{center}
\caption{Comparison with state-of-the-art architectures on ImageNet (mobile setting). }
\label{ev_imagenet}
\end{table*}

\subsection{Architecture Search}

\subsubsection{Implementation Details}

The whole search process consists of 3 stages.
Since we adopt DARTS as the backbone framework,
the search space and network configuration are the same as DARTS at the initial stage (stage 1) except that the number of cells is set to be 5 (this is for acceleration -- we tried the original setting and obtained similar results).
The number of cells increases from 5 to 11 for the intermediate stage (stage 2) and 17 for the final stage (stage 3).
Meanwhile, the size of operation space is set to be 8, 5 and 3 at stage 1, 2 and 3, respectively.

The initial Dropout probability on {\it skip-connect} for the reported results is set to be 0.0, 0.4, 0.7 on CIFAR10 for stage 1, 2 and 3, respectively, and 0.1, 0.2, 0.3 for CIFAR100.
Considering the tradeoff between classification accuracy and computational overhead,
the final discovered cells are restricted to keep at most 2 {\it skip-connect} operations.
Such a setting also guarantees a fair comparison with DARTS and other state-of-the-art approaches.
For each stage, we train the network using a batch size of 96 for 25 epochs,
where in the first 10 epochs only network parameters are tuned
while network parameters and architecture parameters are learned in the rest 15 epochs.
An Adam optimizer with learning rate $\eta = 0.0006$, weight decay 0.001 and momentum $\beta=(0.5, 0.999)$ is adopted for architecture parameters.
GPU memory related hyper-parameters are selected depending on the memory size of the GPU used in the experiments.
For acceleration, we leverage the first order optimization scheme of DARTS to learn the architecture parameters.

\begin{figure*}[t]
\centering
\begin{minipage}{0.49\textwidth}
\subfloat[Stage 1, CIFAR10 Test Err. 2.90\%]{\includegraphics[width=1.0\linewidth]{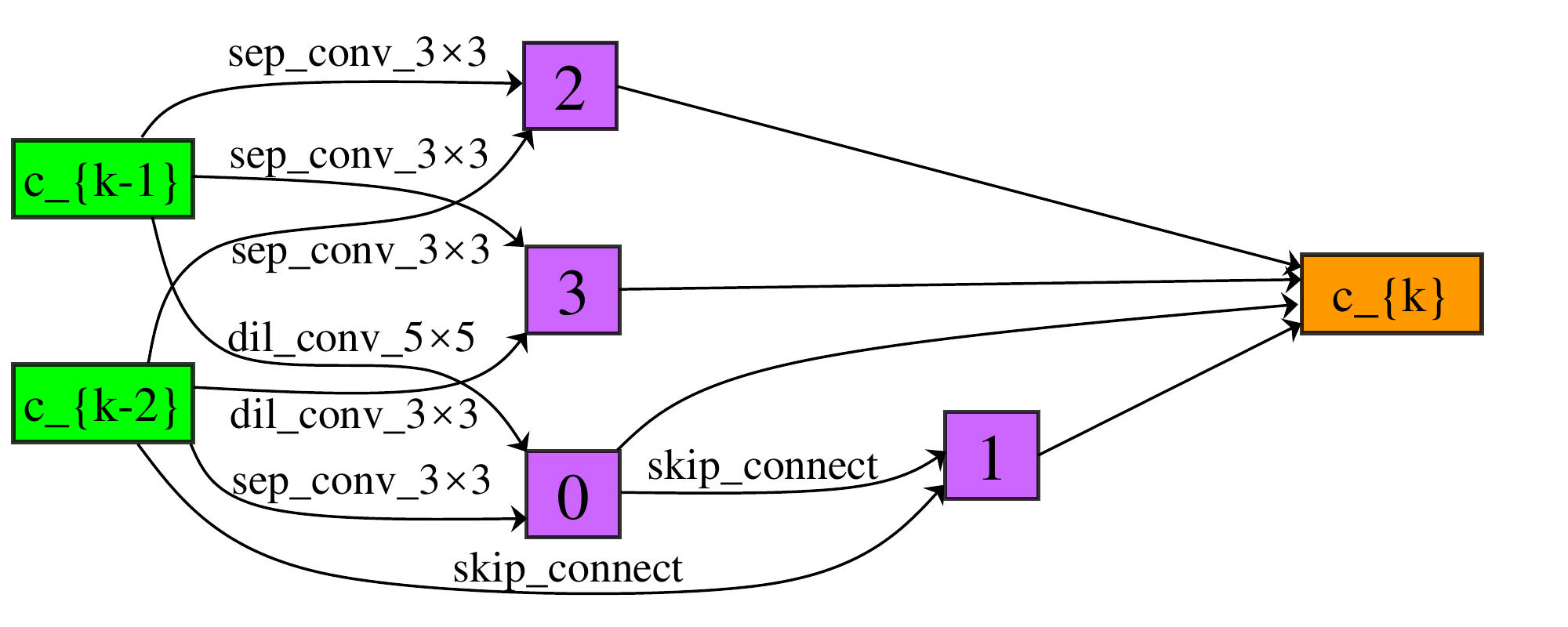}}\label{ncells_s1}\\
\subfloat[Stage 2, CIFAR10 Test Err. 2.82\%]{\includegraphics[width=1.0\linewidth]{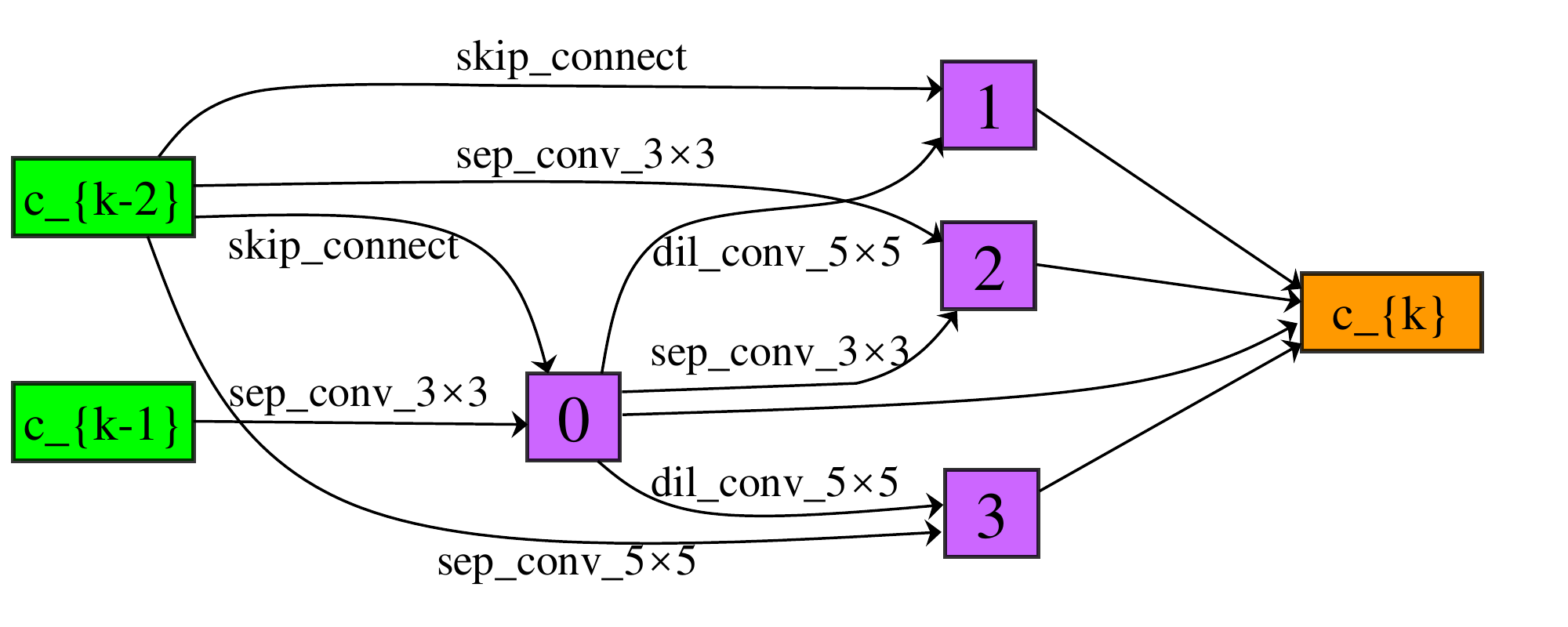}}\label{ncells_s2}
\end{minipage}
\begin{minipage}{0.49\textwidth}
\subfloat[Stage 3, CIFAR10 Test Err. 2.58\%]{\includegraphics[width=1.0\linewidth]{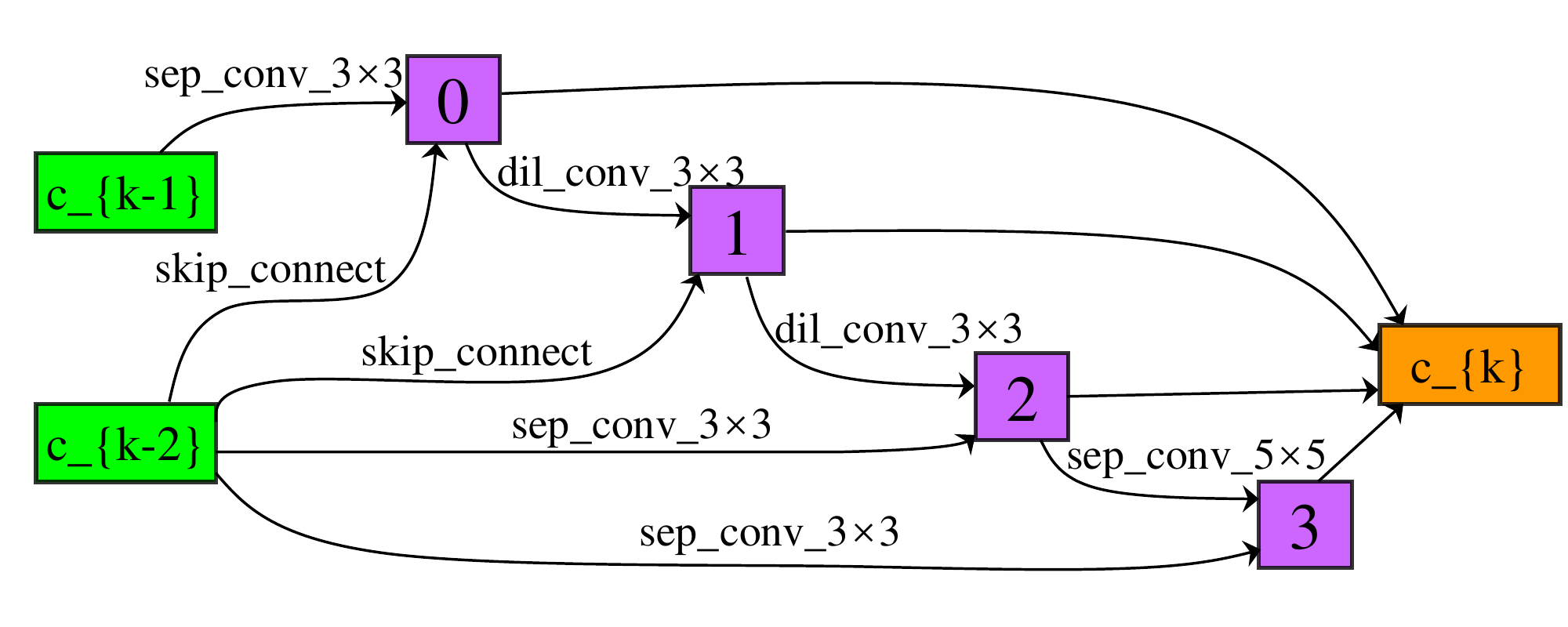}}\label{ncells_s3}\\
\begin{center}
\subfloat[DARTS\_V2, CIFAR10 Test Err. 2.83\%]{\includegraphics[width=1.0\linewidth]{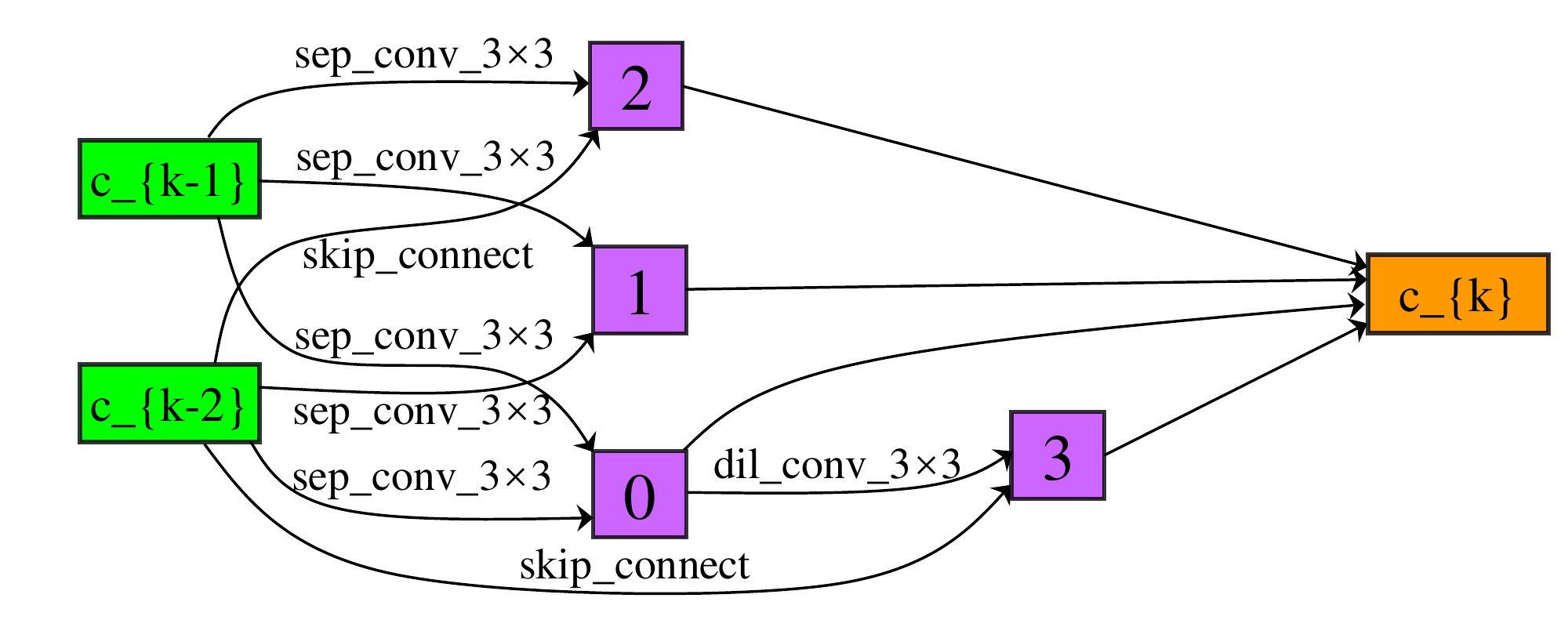}}\label{ncells_dv2}
\end{center}
\end{minipage}
\caption{Normal cells discovered by different search stages of P-DARTS and second order DARTS (DARTS\_V2). The depths of search networks are 5, 11 and 17 cells for stage 1, 2 and 3 of P-DARTS and 8 for DARTS\_V2. When the depth of the search network increases, more deep connections are preserved. Note that the operation on edge $\mathrm{E}_{(0, 1)}$ of stage 1 is a parameter-free {\it skip\_connect}, thus it is strictly not a deep connection.}
\label{ncells}
\end{figure*}

\subsubsection{Search Results}\label{sr}

Architectures discovered by P-DARTS on CIFAR10 tend to preserve more deep connections than the one discovered by DARTS, as shown in Figure~\ref{ncells}(c) and Figure~\ref{ncells}(d).
Moreover, connections in the architecture discovered by P-DARTS cascade for more levels than DARTS, in other words, there are more layers in the cell, making the evaluation network further deeper and achieving better classification performance.

Notably, our method also allows architecture search on CIFAR100 while prior approaches mostly failed.
The evaluation results in Table~\ref{tab_ev_cifar} show that the discovered architecture outperforms those transferred ones.
We also perform architecture search on CIFAR100 with DARTS using the publicly available code
but get an architecture full of {\it skip-connects}, which results in much worse classification performance.

\subsection{Architecture Evaluation}

\subsubsection{Evaluation on CIFAR10 and CIFAR100}

An evaluation network of 20 cells and 36 initial channels is trained from scratch for 600 epochs with batch size 128.
Cutout regularization~\cite{devries2017improved} of length 16, drop-path~\cite{larsson2016fractalnet} of probability 0.3 and auxiliary towers~\cite{szegedy2015going} of weight 0.4 are applied.
A standard SGD optimizer with a weight decay of 0.0003 for CIFAR10 and 0.0005 for CIFAR100 and a momentum of 0.9 is used.
The initial learning rate is 0.025, which is decayed to 0 following the cosine rule.
We further increase the number of initial channels from 36 to 64 to explore the performance limitation of our searched architecture, which is denoted as the large setting.

Evaluation results and comparison with state-of-the-art approaches are summarized in Table~\ref{tab_ev_cifar}.
As demonstrated in Table~\ref{tab_ev_cifar}, P-DARTS achieves a 2.50\% test error on CIFAR10 with a search cost of only 0.3 GPU-days.
To obtain the same performance, AmoebaNet~\cite{real2018regularized} spent four orders of magnitude more computational resources (0.3 GPU-day vs 3150 GPU-days).
Our P-DARTS also outperforms DARTS and SNAS by a large margin.
Notably, architectures discovered by P-DARTS outperform ENAS, the previously most efficient approach, in both classification performance and search cost, while with fewer parameters.

We transfer architectures discovered on CIFAR10 to CIFAR100 for both DARTS and P-DARTS.
The evaluation results show the superiority of P-DARTS.
Furthermore, we also conduct architecture search on CIFAR100.
The discovered architecture outperforms DARTS on both CIFAR10 and CIFAR100.
For a fair comparison, architecture search is also performed on CIFAR100 for DARTS with the publicly released code but get much higher evaluation test error.
An interesting point is that architecture discovered on CIFAR10 outperforms that discovered on CIFAR100 when evaluated on CIFAR10, and vice versa.
Such a phenomenon provides evidence to the existence of dataset bias in NAS.

\subsubsection{Evaluation on ImageNet}

The ILSVRC 2012~\cite{russakovsky2015imagenet} is used to test the transferability of architectures discovered on CIFAR10 and CIFAR100.
We adopt the same network configuration as DARTS, {\em i.e.}, an evaluation network of 14 cells and 48 initial channels.
Each network is trained from scratch for 250 epochs with batch size 1024 on 8 Nvidia Tesla V100 GPUs,
which takes 3 days with our PyTorch~\cite{paszke2017automatic} implementation.
The network parameters are optimized using an SGD optimizer with an initial learning rate of 0.5 (decayed linearly after each epoch), a momentum of 0.9 and a weight decay of $3\times 10^{-5}$.
Additional enhancements including label smoothing~\cite{szegedy2016rethinking} and auxiliary loss tower are applied during training.
Learning rate warmup~\cite{goyal2017accurate} is applied for the first 5 epochs since large batch size and learning rate are adopted.

Evaluation results and comparison with state-of-the-art approaches are summarized in Table~\ref{ev_imagenet}.
Architectures discovered on CIFAR10 and CIFAR100 by P-DARTS outperform DARTS by a large margin in terms of classification performance,
which demonstrates the transfer capability of the discovered architectures.
Notably, P-DARTS achieves lower test error than MnasNet~\cite{tan2018mnasnet} and ProxylessNAS~\cite{cai2018proxylessnas}, whose search space is carefully designed for ImageNet.

\subsection{Diagnostic Experiments} \label{exp_diag}

\subsubsection{Comparison on the Depth of Search Networks}\label{comp_depth}

Since the search process of P-DARTS is split into multiple stages, we extract architectures from each search stage with the same rule.
Architectures of different stages are evaluated to demonstrate their capability for image classification.
The topology of discovered architectures (only normal cells are shown) and their corresponding performance are summarized in Figure~\ref{ncells}.
Additionally, we add the architecture discovered by second order DARTS (DARTS\_V2, 8 cells in the search network) for comparison.

The architecture generated by stage 3 achieves the lowest test error among others, which validates the effectiveness of our scheme.
From Figure~\ref{ncells} we can observe that these architectures share some common edges,
for example {\it sep\_conv\_3$\times$3} at edge $\mathrm{E}_{(c_{k-2}, 2)}$ for stage 1, 2 and 3 and at edge $\mathrm{E}_{(c_{k-1}, 0)}$ for stage 2, 3 and DARTS$\_$V2.
Meanwhile, there are also differences between them, which may be the key factor that affects the capability of these architectures.
Architectures generated by shallow search networks prefer to keep shallow connections,
while with deeper search networks, the discovered architectures start to pick intermediate nodes as input for rear nodes, resulting in cells with deep connections.
This is because it is harder to optimize a deep search network so the algorithm has to explore more paths to find the optimum, which results in more complex and powerful architectures.

\subsubsection{Effectiveness of Search Space Approximation}

The search process takes $\sim$7 hours (0.3 days) on a single Nvidia Tesla P100 GPU with 16GB memory to produce the final architectures.
We collect GPU memory usage data of architecture search process for 3 individual runs, which is shown in Table~\ref{mem_usage}.
The memory usage is stable and out of memory error barely occurs, showing the validity of the search space approximation scheme on memory efficiency.

\begin{table}[htb]
\begin{center}
\begin{tabular}{lccc}
\hline
\multirow{2}{*}{\textbf{Run No.}} & \multicolumn{3}{c}{\textbf{Mem. Usage (GB)}} \\
\cmidrule(lr){2-4}
             & \textbf{Stage 1}  & \textbf{Stage 2} & \textbf{Stage 3} \\
\hline
1  & 9.8 & 14.0 & 14.2 \\
2  & 9.8 & 14.4 & 14.5 \\
3  & 9.8 & 14.2 & 14.3 \\
\hline
\end{tabular}
\end{center}
\caption{Peak GPU memory usage at different stages during three individual runs. The memory limit is 16GB. }
\label{mem_usage}
\end{table}

We perform experiments to demonstrate the effectiveness on improving classification accuracy.
We only perform the final stage of the search process on two different search spaces with the same setting.
The first search space is approximated by the previous search stages and the other is randomly sampled from the full search space.
To obtain a better result for the randomly sampled one, we repeat the whole process for 3 times with different seeds and pick the best one.
The best performance for the randomly sampled search space is 3.43\% test error, which is much worse than 2.58\%, the one obtained by the approximated search space.
Such results reveal the necessity of the search space approximation scheme.

\subsubsection{Effectiveness of Search Space Regularization}

We perform experiments to validate the effectiveness of search space regularization, {\em i.e.}, operation-level Dropout and architecture refinement.
Firstly, experiments are conducted to test the effect of the operation-level Dropout scheme.
Two sets of initial Dropout rates are tested, {\em i.e.}, 0.0, 0.0, 0.0 (without Dropout) and 0.0, 0.3, 0.6 (with Dropout) for stage 1, 2 and 3, respectively.
To eliminate the potential influence of the number of {\it skip-connects}, the comparison is made across multiple values of $M$.

Test errors for architectures discovered without Dropout are 2.93\%, 3.28\% and 3.51\% for $M=2$, $3$ and $4$, respectively.
When searching with Dropout, the corresponding test errors are 2.69\%, 2.84\% and 2.97\%,
significantly outperforming those without Dropout.
According to the experimental results, all 8 operations in the normal cell of the architecture discovered without Dropout are {\it skip-connects} before architecture refinement,
while it is 4 for architecture discovered with Dropout.
The reduction of {\it skip-connect} operations verifies the effectiveness of search space regularization on stabilizing the search process.

During experiments, we observe strong coincidence between the classification performance of an architecture and the number of {\it skip-connect} operations in it.
We conduct a quantitative experiment to verify it.
Architecture refinement is applied to one search process to produce multiple architectures where the number of preserved {\it skip-connect} operations varies from 0 to 4.

The test errors are positively correlated to the number of {\it skip-connects} except for $M=0$, {\em i.e}, 2.78\%, 2.68\%, 2.69\%, 2.84\% and 2.97\% for $M=0$ to $4$, while the parameters count is inversely proportional to it, {\em i.e.}, 4.1M, 3.7M, 3.3M, 3.0M and 2.7M, respectively.
The reason lies in that, with a fixed number of operations in a cell,
the eliminated parameter-free {\it skip-connect} operations are replaced by operations with learnable parameters, {\em e.g.}, {\it convolution},
resulting in a more complex and powerful architecture.

The above observation offers an inspiration for the second kind of search space regularization, architecture refinement, whose capability is validated by the following experiments.
We run another 3 architecture search experiments with initial Dropout rates of 0.0, 0.3 and 0.6 for stage 1, 2 and 3, respectively.
Before architecture refinement, the test error is 2.79 $\pm$ 0.16\% and the evaluated architectures are with 2, 3 and 4 {\it skip-connect} operations in normal cells.
After architecture refinement, all three architectures are with 2 {\it skip-connect} operations in normal cells, resulting in a diminished test error of 2.65 $\pm$ 0.05\%.
The reduction of standard deviation reveals the improvement on the stability for the search process.

\section{Conclusions}

In this work, we propose a progressive version of differentiable architecture search to bridge the depth gap between search and evaluation scenarios.
The core idea is to gradually increase the depth of candidate architectures during the search process.
To alleviate the issues of computational overhead and instability,
we design two practical techniques to approximate and regularize the search process, respectively.
Our approach reports the fastest NAS speed to achieve an error rate of 3\% on CIFAR10, meanwhile achieving superior performance in both the proxy dataset and the target dataset.

Our research defends the importance of depth in differentiable architecture search,
paves a new way of approximation by sacrificing width, {\em i.e.}, the number of operations.
We expect that in the future with more powerful computational resources,
depth is still the dominant factor in exploring the architecture space.

{\small
\bibliographystyle{ieee}
\bibliography{egbib}
}

\end{document}